\pdfoutput=1
\documentclass[11pt]{article}

\usepackage[preprint]{acl}

\usepackage{times}
\usepackage{latexsym}

\usepackage[T1]{fontenc}
\usepackage[utf8]{inputenc}

\usepackage{microtype}

\usepackage{inconsolata}

\usepackage{graphicx}
\usepackage{enumitem}
\usepackage{algorithm}
\usepackage{algpseudocode}
\usepackage[normalem]{ulem}

\usepackage[utf8]{inputenc}    % 编码
\usepackage[T1]{fontenc}       % 字体编码
\usepackage{amsmath, amsfonts}
\usepackage{nicefrac}
\usepackage{microtype}

\PassOptionsToPackage{table}{xcolor}
\usepackage{booktabs}
\usepackage{colortbl}  
\usepackage{multirow}
\usepackage{adjustbox}
\usepackage{hyperref}
\usepackage{tabularx}
\usepackage{hhline}
\usepackage{makecell}
\usepackage{arydshln}
\usepackage{array}
\usepackage{colortbl}         
\usepackage{wrapfig}
\usepackage{graphicx}
\usepackage{algorithm}
\usepackage{hyperref}
\usepackage{url}
\usepackage{caption}

\usepackage[utf8]{inputenc} % allow utf-8 input
\usepackage[T1]{fontenc}    % use 8-bit T1 fonts
\usepackage{hyperref}       % hyperlinks
\usepackage{url}            % simple URL typesetting
\usepackage{booktabs}       % professional-quality tables
\usepackage{amsfonts}       % blackboard math symbols
\usepackage{nicefrac}       % compact symbols for 1/2, etc.
\usepackage{microtype}      % microtypography

\usepackage{placeins}
\definecolor{light_origin}{RGB}{207, 117, 58}
\definecolor{cell_purple}{RGB}{76, 141, 187}
\definecolor{small_blue}{RGB}{27, 63, 125}
\definecolor{small_origin}{RGB}{94, 41, 16}
\definecolor{mypink}{HTML}{D92879}

\definecolor{light_purple}{RGB}{235,236,242}
\definecolor{light_blue}{RGB}{244,249,254}
\newcolumntype{B}{>{\columncolor{blue!4}}c}
\newcolumntype{d}{>{\columncolor{brown!4}}c}
\newcolumntype{q}{>{\columncolor{green!2}}c}
\newcolumntype{R}{>{\columncolor{red!4}}c}     % 浅红色列
\newcolumntype{P}{>{\columncolor{purple!4}}c}   % 浅紫色列
\newcolumntype{Y}{>{\columncolor{yellow!4}}c}   % 浅黄色列

\definecolor{bgcolor}{RGB}{242, 243, 245} % This is defined but not used in the modified table for row backgrounds
\makeatletter
\def\adl@drawiv#1#2#3{%
        \hskip.5\tabcolsep
        \xleaders#3{#2.5\@tempdimb #1{1}#2.5\@tempdimb}%
                #2\z@ plus1fil minus1fil\relax
        \hskip.5\tabcolsep}
\newcommand{\cdashlinelr}[1]{%
  \noalign{\vskip\aboverulesep
            \global\let\@dashdrawstore\adl@draw
            \global\let\adl@draw\adl@drawiv}
  \cdashline{#1}
  \noalign{\global\let\adl@draw\@dashdrawstore
            \vskip\belowrulesep}}

\title{Mind the Generation Process: Fine-Grained Confidence Estimation During LLM Generation}

\author{Jinyi Han\textsuperscript{\rm $\heartsuit$},
Tingyun li\textsuperscript{\rm $\diamondsuit$}, 
Shisong Chen\textsuperscript{\rm $\heartsuit$}, 
Jie shi\textsuperscript{\rm $\spadesuit$},
Xinyi Wang \textsuperscript{\rm $\diamondsuit$}, 
Guanglei Yue \textsuperscript{\rm $\diamondsuit$}, \\
\bf Jiaqing Liang \textsuperscript{\rm $\diamondsuit$}, 
Xin Lin \textsuperscript{\rm $\heartsuit$}, 
Liqian Wen \textsuperscript{\rm $\clubsuit$}, 
Zulong Chen \textsuperscript{\rm $\clubsuit$}, 
Yanghua Xiao\textsuperscript{\rm $\spadesuit$} \thanks{~~Corresponding authors}, \\
\textsuperscript{\rm $\heartsuit$}Shanghai Institute of Artificial Intelligence for Education, East China Normal University\\
\textsuperscript{\rm $\diamondsuit$}School of Data Science, Fudan University\\
\textsuperscript{\rm $\spadesuit$}College of Computer Science and Artificial Intelligence, Fudan University\\
\textsuperscript{\rm $\clubsuit$} Alibaba \\
\texttt{\{jinyihan099, litinyun0715@gmail.com\}} \\
}

\begin{document}
\maketitle

\begin{abstract}
While large language models (LLMs) have demonstrated remarkable performance across diverse tasks, they fundamentally lack self-awareness and frequently exhibit overconfidence, assigning high confidence scores to incorrect predictions. Accurate confidence estimation is therefore critical for enhancing the trustworthiness and reliability of LLM-generated outputs. However, existing approaches suffer from coarse-grained scoring mechanisms that fail to provide fine-grained, continuous confidence estimates throughout the generation process. To address these limitations, we introduce FineCE, a novel confidence estimation method that delivers accurate, fine-grained confidence scores during text generation. Specifically, we first develop a comprehensive pipeline for constructing training data that effectively captures the underlying probabilistic distribution of LLM responses, and then train a model to predict confidence scores for arbitrary text sequences in a supervised manner. Furthermore, we propose a Backward Confidence Integration (BCI) strategy that leverages information from the subsequent text to enhance confidence estimation for the current sequence during inference. We also introduce three strategies for identifying optimal positions to perform confidence estimation within the generation process. Extensive experiments on multiple benchmark datasets demonstrate that FineCE consistently outperforms existing classical confidence estimation methods. Our code and all baselines used in the paper are available on GitHub \footnote{https://github.com/JinyiHan99/FineCE}.
\end{abstract}

\section{Introduction}
Self-awareness, as a core metacognitive ability, plays a crucial role in both human cognition and the advancement of large-scale AI systems \cite{dewey1986experience, kuhl2012action}. For humans, it enables reflective thinking and error monitoring. Similarly, for large language models (LLMs), it supports output evaluation and self-correction, which is critical for handling complex reasoning tasks \cite{tong-etal-2024-llms, xie2025teaching}. Confidence estimation has emerged as a promising approach, enabling models to assess the reliability of their own generations \cite{zhou2023navigating, xiong2023can, Branwen}.
\begin{figure}[t]
  \centering
  \includegraphics[width=.99\linewidth]{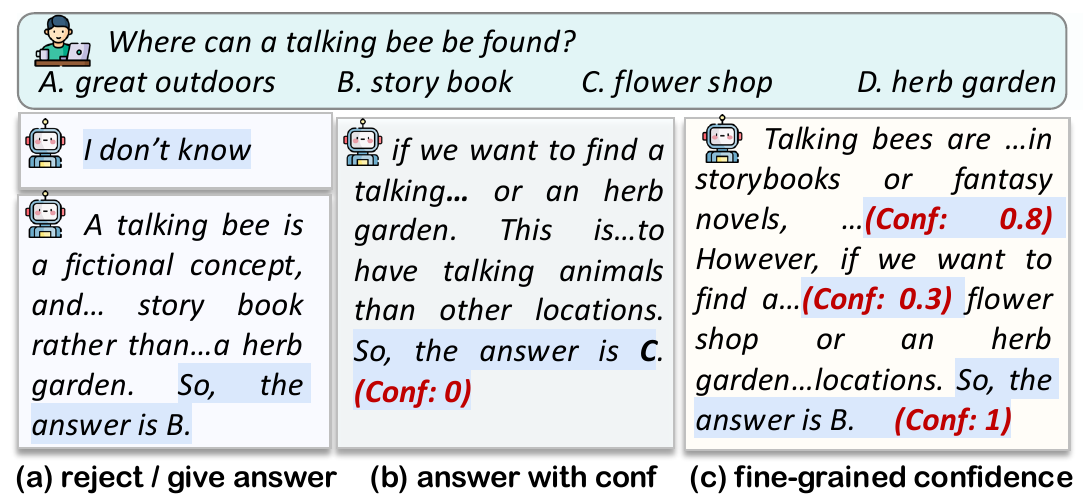}
  \caption{\footnotesize{The difference between our proposed FineCE and existing confidence estimation methods. \textbf{(a):} LLMs either generate an answer when the query is within their knowledge scope or refuse to answer if it falls beyond their capabilities. \textbf{(b):} The model assigns a single confidence score after the entire answer is generated. \textbf{(c):} Our proposed method, FineCE, provides the fine-grained confidence scores for any given text sequence throughout the generation process.}}
  \label{intro}
\end{figure}

However, existing confidence estimation methods for LLMs remain limited by their \textbf{coarse-grained} scoring and \textbf{narrow perspective}, failing to provide reliable and continuous confidence signals. Broadly, these works are categorized into question-oriented and outcome-oriented paradigms. \textit{\textbf{Question-oriented}} methods aim to constrain LLMs to answer only questions within their domain of knowledge, allowing the model to give up responding when uncertain \cite{Zhang2023RTuningIL}. 
When faced with ambiguous or challenging questions, LLMs often decline to answer such questions directly \cite{Kadavath2022LanguageM}, rather than attempting to infer a potential answer from the available context. While this conservative method helps prevent the model from generating incorrect answers, it also significantly limits the utility of LLMs in open-ended tasks. \textit{\textbf{Outcome-oriented}} methods require LLMs to evaluate the quality of their generated answers after completing the generation process \cite{Zhang2024CalibratingTC, Zhao2024FactandReflectionI, Kuhn2023SemanticUL, AbbasiYadkori2024ToBO}. However, relying solely on a single confidence score at the end of the generation is insufficient to capture the model’s certainty throughout the entire reasoning trajectory. A high final confidence score does not indicate that the intermediate steps are completely accurate \cite{jiaolearning}. Figure \ref{intro} highlights the key differences between these two confidence estimation paradigms.
\label{benefits}

Therefore, it is essential to develop \textbf{fine-grained confidence estimation} methods that provide accurate confidence scores for the intermediate steps during generation.
This enables \textbf{early prediction} of whether the model is likely to produce a correct final answer, without having to wait for the full response. In addition, intermediate confidence scores serve as \textbf{supervisory signals} for LLMs with deep thinking capabilities, such as O1\footnote{\url{https://openai.com/openai-o1-contributions}} and R1 \cite{guo2025deepseek}. These signals inform the model’s decision-making during generation, determining whether to proceed with the current trajectory or to revise earlier outputs. Furthermore, questions that consistently lead to low confidence scores expose \textbf{underlying weaknesses} in the model, offering actionable insights for targeted improvements.

Implementing fine-grained confidence estimation in LLMs is non-trivial and presents three major challenges. \textit{\textbf{(Task Learning:)}} \textit{In the absence of explicit confidence annotations, how can we teach LLMs to express fine-grained confidence? } LLMs are not inherently equipped with such capability \cite{Tian2023JustAF}.
% Although they possess internal state representations, such as logits and strong instruction-following capability, these alone are insufficient for producing reliable fine-grained confidence estimates \cite{Su2024UnsupervisedRH, Chen2024SelfIESO, Branwen}.
Dedicated and task-specific supervised training is necessary. However, constructing supervisory data for this task poses a significant challenge. A key difficulty lies in the fact that distilling confidence scores from other advanced models is impractical, as the uncertainty captured by these models does not necessarily reflect that of the learner model itself.% Additionally, since LLMs typically generate unstructured, free-form text, accurately assigning confidence scores to arbitrary outputs remains inherently difficult.
\textit{\textbf{(Effectiveness:)}} \textit{How to provide accurate and unbiased confidence estimate for the current text?} During generation, LLMs predict each token sequentially without access to future content. Relying solely on confidence scores derived from the current partial output easily introduces bias and miscalibration.
\textit{\textbf{(Efficiency:)}} \textit{What are the optimal positions for confidence estimation?} Estimating confidence after every generated token is often unnecessary and computationally inefficient. Instead, it is crucial to identify key positions during generation where confidence estimation has the greatest impact and provides the most value.

In this paper, we introduce FineCE, a fine-grained confidence estimation method for LLMs via supervised learning. Specifically, to capture the distributional uncertainty inherent in an LLM, we design a complete data construction pipeline based on Monte Carlo Sampling. Additionally, we introduce a Backward Confidence Integration (BCI) strategy at the inference stage, which further refines the confidence estimation for current predictions by utilizing uncertainty information from subsequently generated tokens. To better balance the trade-off between confidence estimation performance and computational efficiency, we propose three strategies to identify optimal positions within the generation process for performing confidence estimation.

% FineCE significantly outperforms existing confidence estimation methods across multiple metrics.
% % and widely-used open-source LLMs. 
% Notably,
Experiments demonstrate that FineCE can reliably estimate the likelihood of a correct final answer as early as one-third into the generation process, offering strong early-stage confidence signals. To further validate its effectiveness, we apply FineCE to a downstream task using a confidence-based filtering strategy that retains only responses exceeding a predefined threshold. This strategy leads to a substantial 39.5\% improvement in accuracy on the GSM8K dataset.

In summary, our contributions are four-fold: 
\begin{itemize}[noitemsep,left=0pt]
 \item We propose FineCE, a fine-grained confidence estimation method that enables accurate prediction of answer correctness during the generation process.
 \item We design a complete pipeline for constructing high-quality training data that effectively captures the distributional uncertainty of LLMs.
 \item We introduce BCI, a novel backward confidence integration strategy that enhances current confidence estimation by incorporating uncertainty information from subsequent texts.
 \item We develop three practical strategies to identify optimal positions for confidence estimation within the generation process.
\end{itemize}

% In this section, we first establish the formal formalization of the task our work aims to solve.
% To address the three core challenges, we introduce the complete process of constructing training data and provide three strategies to assist in determining when to perform confidence estimation. Additionally, we propose a Backward Confidence Integration strategy for the testing phase.

\section{Task Formalization}
The confidence estimation task aims to improve model calibration by aligning predicted probabilities with the likelihood of correct outputs. Here, \textbf{\textit{confidence is defined as the probability that the model’s answer is correct}}.

Formally, LLMs generally generate responses in an auto-regressive manner, predicting the next token sequentially based on the previously generated context. Given an input $x$ and an LLM \(M\), the model generate a sequence of output tokens $y = {t_1, t_2, \cdots, t_n}$, where each token $t_i$ is sampled from the distribution $P_i = \mathcal{P}(\cdot \mid x, t_{<i};M)$, with $t_{<i} = {t_1, \cdots, t_{i-1}}$ and $n$ denoting the total number of generated tokens. Let $\bar{Y}$ denote the ground-truth output. Given any intermediate generation sequence $s$, we define the confidence score as:
\begin{equation}
    Conf_{s} = p( y = \bar{Y} | s, M)
\end{equation}
The confidence score $Conf_s$ of a sequence $s$,  which can be a partial or complete answer, represents the probability that model $M$ generates the correct output $\bar{Y}$, conditioned on $s$. 
Depending on the form of $s$, we categorize the confidence estimation task into the following three variants:
\begin{itemize}[noitemsep,left=0pt]
    \item \textbf{Question-oriented confidence estimation.}
    In this setting, \(s\) contains only the input question, that is, \(s=x\). 
    % The model is required to estimate the confidence of generating a correct answer before producing any tokens.
    \item \textbf{Process-oriented confidence estimation.} \( s \) consists of the input question and a partially generated answer, i.e., \( s = (x, t_{<i}) \), where \( t_{<i} \) is a prefix of the full output sequence \( y \). 
    % This variant allows the model to estimate confidence at any intermediate step during generation.
    \item \textbf{Outcome-oriented confidence estimation.} In this case, \( s \) includes both the input and the complete generated response, that is, \( s = (x, y) \). 
    % The confidence score reflects the model’s belief in the correctness of the final output.
\end{itemize}

This formulation unifies existing confidence estimation settings under a common probabilistic view. It also extends the task to cover all stages of the generation process.
\section{FineCE: Fine-grained Confidence Estimation}
\begin{figure*}[ht]
  \centering
\includegraphics[width=.9\linewidth]{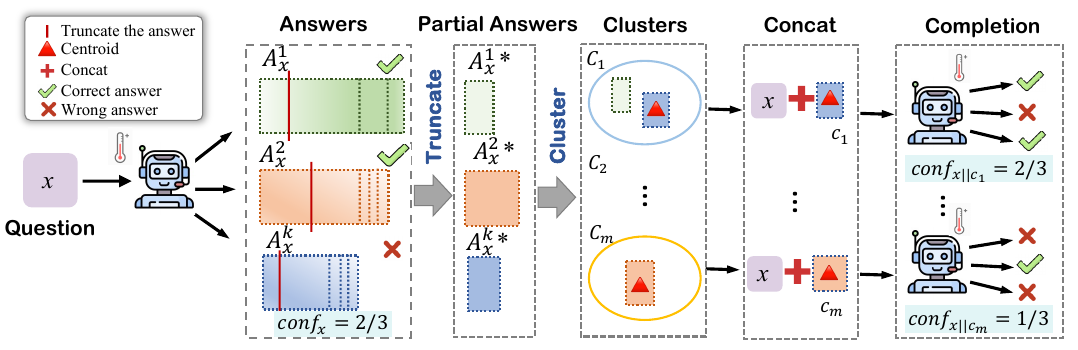}
\caption{\footnotesize{ The construction process of the training dataset. It illustrates the confidence scoring procedures for \textit{Question} and Q\textit{uestion with Partial Answer} using Monte Carlo sampling. For \textit{Question with Answer}, the confidence score is determined based on the correctness of the answer. The complete data construction procedure is detailed in Algorithm \ref{alg:conf_data}.
   }}
\label{fig:pipeline}
\end{figure*}
\subsection{Data Construction}
\label{construction}
% \textbf{Preliminary.} Traditional deep learning approaches for classification fail to capture the model uncertainty. The predictive probabilities provided by the softmax output are frequently misinterpreted as a measure of the model's confidence. However, the model may still be uncertain in its predictions despite producing a high softmax output \cite{gal2016dropout}. 
\textbf{Preliminary.} Traditional classification models struggle to reflect predictive uncertainty, as softmax probabilities are often misinterpreted as confidence scores. A high softmax output does not necessarily indicate that the model is certain about its prediction \cite{gal2016dropout}.
Therefore, to obtain the LLM's inherent real responses probability based on the text $s$, we introduce Monte Carlo Sampling\cite{Li2024NeuroSymbolicDG} and employ the generative LLM $M$ to repeatedly sample $k$ answers $\{A_{s}^{1}, A_{s}^{2}, \cdots, A_{s}^{k}\}$ at high temperature to approximate the probability of generating the correct answer. According to the \textit{Law of Large Numbers}, as $k$ approaches infinity, the sample mean will converge to the true probability of the model generating the correct answer.\\
% \\The key advantage of Monte Carlo sampling is its ability to provide an unbiased and statistically reliable estimate of the model’s performance.
\textbf{Overall Pipeline.}
In our work, the input text sequence $s$ includes three distinct types: \textit{Question},\textit{ Question with Partial Answer} and \textit{Question with Answer}. The confidence score $Conf_s$ is calculated as the accuracy ratio of \(k\) generated answers compared to a reference or golden answer \(\bar{Y}\), which is defined as follows:
\begin{equation}
     Conf_s = \frac{{\textstyle \sum_{i=1}^{k}} \mathbf{I}(A_{s}^{i}=\bar{y}_s)}{k},
\end{equation}
where $A_s^i$ is the $i$th sampling answer generated based on sequence $s$, and $\bar{y}_s$  is the ground-truth answer. The indicator function $\mathbf{I}$ returns 1 when the answer matches and 0 otherwise.
\noindent
\paragraph{Confidence score for \textit{Question}.} 
For each input question $x$, we first generate $k$ diverse complete answers $\{A_x^{1},A_x^{2},\cdots,A_x^{k}\}$ from the model $M$ using a high-temperature sampling strategy. Here, $A_x^{i}$ represents the $i$th response conditioned on input $x$. The confidence score for $x$ is calculated according to Equation 2.

\paragraph{Confidence score for \textit{Question with Partial Answer}.} To construct training data for confidence estimation on partial answers, we apply a truncation procedure to each complete answer $A_x^{i}$, yielding a sequence of partial answer fragments. Each fragment is then concatenated with the original question $x$ and fed into the model to generate multiple completions. These completions are subsequently used to estimate the confidence score associated with the partial answer.

We leverage an intrinsic property of LLMs to reduce the computational overhead associated with constructing training datasets. Specifically, when processing inputs with identical prefixes, their internal contextual representations tend to converge, resulting in highly similar conditional probability distributions for subsequent generations \cite{latentconvergen}. 
% This phenomenon leads to increased output consistency, which strengthens as the shared prefix lengthens. 

Based on this observation, we propose a \textit{progressive data construction pipeline}. Starting with an initial set of $k$ partially completed answer fragments obtained via truncation, we first perform semantic clustering to group these fragments into $m$ clusters, where $1 \le m \le k$. Each cluster contains semantically similar fragments. We then select a centroid fragment from each cluster to serve as its representative. Each selected representative is then concatenated with the original question to generate $k$ new complete answer trajectories through Monte Carlo sampling, which is facilitates the estimation of a confidence score for each representative. From the sampled trajectories, we identify a semantically representative answer and apply another truncation operation to obtain a new partial answer. 

This process is iteratively repeated, with each iteration yielding  new set of partial answers along with the confidence estimates. The total number of truncation is limited to a maximum of $\mathcal{T}$.
\paragraph{Confidence score for \textit{Question with Answer}.} Upon completion of the process described above, we obtain a diverse set of partial answers, each associated with a corresponding confidence estimate. Simultaneously, each Monte Carlo sampling step yields a complete answer to the input question $x$. If a sampled answer matches the ground truth, it is assigned a confidence score of 1.0; otherwise, it receives a score of 0.0. 

The overall training data construction pipeline is illustrated in Figure \ref{fig:pipeline} and detailed in Algorithm 1. The formats of three data types shown in Figure \ref{fig:training_data_format}.\\
\textbf{Complexity Analysis.} 
The primary cost in constructing the training dataset arises from the number of forward passes required during Monte Carlo sampling. Without any optimization, generating three types of confidence estimates for each problem instance leads to an exponential growth in overall generation cost. This process can be viewed as maintaining a full k-ary tree of depth $\mathcal{T}+1$, resulting in a total of $\sum_{i=1}^{\mathcal{T}+1}k^{i}$ model inferences. To reduce complexity, clustering based on semantic similarity can be performed among sibling nodes at each hierarchical level. The generation cost is reduced to $k\sum_{i=0}^{\mathcal{T}}m^{i}$. Here, instead of first clustering the $k$ generated candidates and then selecting the centroid of each cluster, we perform truncation by directly selecting a semantically representative candidate from the $k$ answers at each step, from the 2nd to the $\mathcal{T}$-th. This strategy significantly reduces the total generation cost to $k(1 + m\mathcal{T})$. As a result, in our work, the overall complexity of constructing the training data is \textbf{reduced from exponential to linear with respect to $\mathcal{T}$}.

\subsection{Training Technique} To enhance the confidence estimation capability of LLMs, we explore two distinct training techniques, including the Additional Value Head and Instruction Fine-Tuning (IFT) \cite{Ouyang2022TrainingLM}. The additional value head skill reformulates confidence estimation as a multi-classification task, enabling token-level confidence predictions across the generated sequence. In contrast, IFT leverages the model's natural language generation capabilities to produce confidence estimates in a more interpretable format and human-readable format. In the Figure \ref{training_skill}, we provide a comprehensive comparison of these two technique in our proposed task. In this work, FineCE adopts the IFT training paradigm. 

\subsection{Identify the Calibration Position} 
FineCE introduces fine-grained confidence estimation for LLMs. Calibrating confidence after each token generation is impractical due to computational costs.
% We propose three strategies to identify optimal positions for confidence estimation during the generation process.
To reduce the computational overhead of token-wise confidence calibration, FineCE introduces three strategies to selectively perform confidence estimation during generation.

\textbf{Paragraph-End Calibration} conducts estimation at natural linguistic boundaries, such as paragraph ends. It maintains semantic coherence with minimal disruption to the generation flow.

% This strategy performs confidence estimation at natural sentence boundaries, leveraging linguistic breaks in the generation process. By calibrating at paragraph endpoints, it minimizes the disruption to the generation flow while preserving semantic coherence and contextual integrity.
\textbf{Periodic Calibration} performs estimation at fixed token intervals (e.g., every 50 tokens). This regular, interval-based strategy offers a deterministic mechanism for confidence monitoring, ensuring consistent quality assessment across the entire generated sequence.

\textbf{Entropy-based Calibration} triggers estimation when the model’s output entropy exceeds a predefined threshold. While entropy reflects uncertainty, it alone is not sufficient for accurate confidence prediction. The calibration is more meaningful and reliable when entropy values are higher.

% We aim to identify an effective strategy and establish basic guidelines for selecting appropriate confidence estimation positions in different generation scenarios.
\subsection{Backward Confidence Integration (BCI)}
Existing confidence estimation methods rely solely on local features while overlooking the global context, resulting in incomplete or biased estimation. However, training data construction typically adopts a backward evaluation paradigm, labeling intermediate steps based on the correctness of the final answer \cite{tree-of-thought, ICLR2025_MCTS}. Yet, this valuable supervision signal is rarely exploited during inference. Therefore, to further revise either excessively high or low confidence level and mitigate output confidence bias, we propose Backward Confidence Integration (BCI). It extends the backward evaluation principle from training to inference.
% BCI's core innovation lies in incorporating future context into current confidence estimation. By leveraging information from later reasoning steps, BCI enables dynamic refinement of earlier confidence estimates. This global perspective mitigates local estimation bias and produces more stable, context-aware confidence scores across the reasoning trajectory.
 % \cite{tree-of-thought, ICLR2025_MCTS}
% Current confidence estimation methods primarily rely on local features, ignoring the broader reasoning context. In multi-step reasoning, the reliability of each step is influenced by surrounding steps, making local estimates insufficient to capture true confidence.

% We usually assess the correctness of intermediate steps based on the correctness of the final answer. This backward paradigm is widely adopted in training data construction. BCI naturally extends this idea to the inference stage. Rather than relying solely on the final output for confidence estimation, BCI enables dynamic correction of earlier steps by incorporating information from later context.

% To further revise either excessively high or low confidence level and mitigate output confidence bias, we introduce the Backward Confidence Integration strategy. This strategy incorporates the future context into the current confidence estimation, enabling a more globally informed and stable estimation. 

Formally, for a generated text sequence, $Conf_{s_{j}}$ denotes the initial confidence estimation at the $j$th calibration position in a generated sequence. The adjusted confidence score $Conf_{s_{h}}^{'}$ is computed recursively for positions $h \in (j,j+d)$, which is defined as:\\
\begin{equation}
{\scriptstyle
Conf'_{s_j} =
\begin{cases} 
\alpha Conf_{s_j} + (1-\alpha) \frac{1}{w} \sum_{b=1}^{w} Conf'_{s_{h+1}^b},\\ \hspace{11em}h < j + d\\
Conf_{s_h}, h = j + d
\end{cases}
}
\end{equation}
Here, $\alpha \in [0,1]$ is the revision coefficient balancing the original local confidence and the influence of future context. A smaller $\alpha$ places placing more weight on future text. The parameters $w$ defines the number of sampled generation paths (integration width), and $d$ specifies how many future positions are considered (integration depth). $Conf_{s_{h}^{b}}$ denotes the adjusted confidence at the $h$th calibration position in the $b$th sample. By recursively incorporating backward signals from future steps, it provides a more globally accurate estimation of confidence for each calibration position. 
% An illustrative example is provided in Figure \ref{BCI}.

\section{Experiments}
% We conduct extensive experiments to verify the effectiveness of FineCE, focusing on confidence estimation performance,and analyze the effectiveness of the BCI strategy.  \\

\subsection{Experiment Setting}
\textbf{Dataset and Metrics.} We evaluate the performance of confidence estimation across six datasets including \textit{GSM8K}\cite{cobbe2021gsm8k}, \textit{TriviaQA}\cite{Joshi2017TriviaQAAL}, \textit{CommonsenseQA}(CSQA; \cite{talmor2018commonsenseqa}), \textit{AIME24}\footnote{https://huggingface.co/datasets/math-ai/aime24}, \textit{MMLU} \cite{MMLU}, and \textit{NQ-Open} \cite{kwiatkowski-etal-2019-natural}. 
% We adopt several widely used metrics including Expected Calibration Error (ECE), Receiver Operating Characteristic Curve (AUROC) and Accuracy (ACC). 

We adopt several widely used metrics including Expected Calibration Error (ECE), Receiver Operating Characteristic Curve (AUROC) and Accuracy (ACC). \\
\textbf{Models and Baselines.} We employ three widely-used open-source models, including Llama2-13B \cite{Touvron2023Llama2O}, Llama3.1-8B \cite{dubey2024llama} and Qwen2.5-7B \cite{yang2024qwen2}. The baselines we used in this paper include the following three types:
1) \textbf{Question-oriented:} {P(IK)} \cite{Kadavath2022LanguageM}; 2) \textbf{Outcome-oriented:} {{First-Prob}}, {SuC} \cite{lin2022teaching}, {Verbalized Porb} (Verb \cite{Tian2023JustAF}) {Semantic Uncertainty} (SE, \cite{kuhn2023semantic}); 3) \textbf{Step-wise estimation:} {Multi-Step} (MS; \cite{xiong2023can}), {LECO} \cite{Yao2024LearningFC}.
% ECE evaluates how well a model’s confidence estimates aligns with its actual accuracy. AUROC gauges the model’s capability to assign higher confidence to correct predictions and lower confidence to incorrect ones, aiming to determine if confidence scores can effectively distinguish between correct and incorrect predictions. Besides, we use \textit{Accuracy (ACC)} to evaluate the accuracy performance of these baselines.

% Further details about datasets, baselines, implementations (including all prompts used in this paper, important parameters, and platforms) can be found in Appendix \ref{exp_details}. 
% In addition, in Appendix \ref{further_discuss}, we also present an in-depth discussion on FineCE’s generalization ability, its dependence on training data, the impact of training strategies, and its performance on highly open-ended questions. 在正文中，我们先讨论了西面四个问题
Comprehensive experimental details, including dataset baseline introduction, prompts used, key hyperparameters, and computational platforms, are provided in Appendix. Beyond the core results presented in the main text, we conduct additional analyses to address four critical questions regarding FineCE's practical applicability: (1) generalization ability across different domains, (2) sensitivity to training data, (3) impact of different training strategies, and (4) performance on highly open-ended questions of FineCE. 
% These supplementary analyses are detailed in Appendix.

\subsection{Main Results and Analysis}
\textbf{RQ1: How does FineCE perform compared with baselines?} In this experiment, to ensure fair comparison, we fix the parameters $w$ and $b$ in FineCE to 0, eliminating the computational advantage of BCI, thereby aligning inference costs with baseline methods. The overall results are shown in Table \ref{tab:process_orient} and Table \ref{tab:qa_oriented}.

\renewcommand{\arraystretch}{0.4} 
\setlength{\belowcaptionskip}{2pt} % 表格标题与正文之间的间距
\vspace{-2em} % 减少间距
\begin{table*}[ht]
\renewcommand{\familydefault}{\rmdefault}
  \centering
  \caption{Confidence estimation results throughout the generation process. \(z\) is total number of paragraphs in an answer. $p(1)$ and $p(z-1)$ represent the confidence estimates for the first and the penultimate paragraphs of the generated answer, respectively.   }
   \captionsetup{skip=2pt} 
% \small
\begin{adjustbox}{width=.85\textwidth}
    \begin{tabular}{cr r BBB ddd qqq}
    \toprule
     &\multirow{2}{*}{\textbf{Pos}} &\multirow{2}{*}{\textbf{Metrics}} &\multicolumn{3}{c}{\textbf{Llama2-13B}} &\multicolumn{3}{c}{\textbf{Llama3.1-8B}} &\multicolumn{3}{c}{\textbf{Qwen2.5-7B}}\\
     % \specialrule{0em}{1pt}{1pt}
     % \cline{4-6} \cline{8-10} \cline{12-14} 
       
       & & & \cellcolor{white}{MS}	& \cellcolor{white}LECO & \cellcolor{white}FineCE & \cellcolor{white}MS	& \cellcolor{white}LECO & \cellcolor{white}FineCE & \cellcolor{white}MS	& \cellcolor{white}LECO & \cellcolor{white}FineCE  \\ 
       % \specialrule{0em}{1pt}{1pt}
     \midrule
     \multirow{6}{*}{\rotatebox{90}{\textbf{GSM8K}}} &\multirow{2}{*}{$p(1)$} 
     &AUROC$\uparrow$	&55.6 	&60.5	&\textbf{73.8}	&60.8 &62.2 &\textbf{66.2} &64.7 &64.4 &\textbf{66.8}\\
     &&ECE$\downarrow$	&23.5	&19.2 	&\textbf{9.3}	&27.4 &21.1 &\textbf{15.7} &23.6 &21.1 &\textbf{14.1}\\
 
     \cdashlinelr{2-12}
     &\multirow{2}{*}{$p(z-1)$} 
    &AUROC$\uparrow$	&57.3 	&59.5	&\textbf{77.7}	&62.3 &64.7 &\textbf{69.4} &63.8 &65.3 &\textbf{65.3}\\
    &&ECE$\downarrow$	&22.8	&21.3  & \textbf{8.4} &29.7 &23.7 &\textbf{17.3} &25.2 &20.4 &\textbf{14.4} \\
    \cdashlinelr{2-12}
     &\multirow{2}{*}{$AVG$}      &AUROC$\uparrow$	&57.1 	&61.1	&\textbf{78.1}	&62.4 &68.2 &\textbf{72.7}  &67.2 &64.1 &\textbf{76.4} \\
     &&ECE$\downarrow$ &21.1	&19.6 	&\textbf{6.7}	&28.3 &19.2 &\textbf{12.3}  &19.2 &20.1 &\textbf{10.7}  \\

     \midrule
     \multirow{6}{*}{\rotatebox{90}{\textbf{CSQA}}} &\multirow{2}{*}{$p(1)$}  &AUROC$\uparrow$	&54.6 	&57.1	&\textbf{66.2}	&61.0 &63.1 &\textbf{66.3} &63.9 &62.0 &\textbf{68.1} \\ 
     & &ECE$\downarrow$ &24.8	&23.8 	&\textbf{18.3}	&29.4 &22.4 &\textbf{16.6}  &27.6 &19.2 &\textbf{17.3} \\
    
     \cdashlinelr{2-12}
    &\multirow{2}{*}{$p(z-1)$} 
     &AUROC$\uparrow$	&53.2 	&56.0	&\textbf{69.3}	&57.2 &62.9 &\textbf{67.5} &62.0 &63.9 &\textbf{68.2}  \\
    &&ECE$\downarrow$	&26.9	&25.7 	&\textbf{16.2}	&33.0 &26.3 &\textbf{17.9}  &24.4 &20.8 &\textbf{17.1} \\
   
     \cdashlinelr{2-12}
     &\multirow{2}{*}{$AVG$}   &AUROC$\uparrow$	&58.6 	&59.6	&\textbf{71.3} &59.3 &65.0 &\textbf{71.1} &65.5 &65.3 &\textbf{73.2}  \\
     &&ECE$\downarrow$ &23.1	&21.4 	&\textbf{11.7}	&29.3 &23.1 &\textbf{13.3} &25.0 &17.6 &\textbf{14.7} \\
   
     \midrule
      \multirow{6}{*}{\rotatebox{90}{\textbf{TriviaQA}}} &\multirow{2}{*}{$p(1)$} 
       &AUROC$\uparrow$	&56.1 	&53.4	&\textbf{70.8}	&63.4 &60.7 &\textbf{69.2} &61.9 &62.1 &\textbf{67.4} \\
      &&ECE$\downarrow$ &22.2	&26.8 	&\textbf{14.5}	&27.9 &21.4 &\textbf{18.7} &26.4 &22.7 &\textbf{19.3} \\
     
      \cdashlinelr{2-12}
      &\multirow{2}{*}{$p(z-1)$} 
        &AUROC$\uparrow$	&56.4	&58.3	&\textbf{74.2}	&62.0 &63.4 &\textbf{67.7} &59.4 &64.4 &\textbf{71.1} \\
      &&ECE$\downarrow$	&25.6	&27.3	&\textbf{15.0}	&26.3 &20.9 &\textbf{20.3} &30.2 &23.4 &\textbf{17.5} \\
    
      \cdashlinelr{2-12}
&\multirow{2}{*}{$AVG$} 
&AUROC$\uparrow$	&57.2	&58.1	&\textbf{76.1}	&63.7 &62.6 &\textbf{73.3} &63.2 &64.0 &\textbf{73.9}\\ 
&&ECE$\downarrow$	&22.8	&25.5	&\textbf{11.3} &25.1 &19.3 &\textbf{14.2} &25.3 &20.2 &\textbf{13.4}\\

      \midrule
      \multirow{6}{*}{\rotatebox{90}{\textbf{AIME24}}} &\multirow{2}{*}{$p(1)$} 
         &AUROC$\uparrow$	&21.4 &56.3 &\textbf{68.4} &16.2 &63.4 &\textbf{69.8} &25.3 &64.1 &\textbf{74.1} \\
      &&ECE$\downarrow$ &57.4 &37.4 &\textbf{19.3} &60.3 &31.2 &\textbf{21.5} &64.3 &33.7 &\textbf{22.4}  \\
   
      \cdashlinelr{2-12}
      &\multirow{2}{*}{$p(z-1)$} 
      &AUROC$\uparrow$	&25.4 &59.4 &\textbf{71.3} &25.3 &66.3 &\textbf{68.4} &11.6 &65.2 &\textbf{76.2}  \\
      &&ECE$\downarrow$	&64.3 &34.3 &\textbf{22.4} &57.2 &29.4 &\textbf{23.5} &76.8 &30.2 &\textbf{21.3}  \\
  
      \cdashlinelr{2-12}
&\multirow{2}{*}{$AVG$} 
&AUROC$\uparrow$	&22.7 &56.3 &\textbf{76.0} &19.5 &64.1 &\textbf{71.3} &30.3 &64.0 &\textbf{79.2}  \\
&&ECE$\downarrow$	&59.2 &33.8 &\textbf{16.5} &55.4 &30.8 &\textbf{20.4} &72.3 &29.6 &\textbf{18.3}\\
 
\midrule
      \multirow{6}{*}{\rotatebox{90}{\textbf{MMLU}}} &\multirow{2}{*}{$p(1)$} 
        &AUROC$\uparrow$	&57.4 &61.3 &\textbf{74.3} &53.1 &59.2 &\textbf{70.3} &54.1 &60.3 &\textbf{70.2}  \\
      &&ECE$\downarrow$ &27.6 &26.2 &\textbf{20.1} &30.3 &27.8 &\textbf{20.2} &32.9 &30.3 &\textbf{22.4} \\
    
      \cdashlinelr{2-12}
      &\multirow{2}{*}{$p(z-1)$} 
      &AUROC$\uparrow$	&59.3 &62.5 &\textbf{71.8} &56.4 &61.3 &\textbf{73.1} &52.6 &57.4 &\textbf{71.3} \\
      &&ECE$\downarrow$ &29.4 &28.1 &\textbf{18.9} &33.6 &29.3 &\textbf{17.3} &33.4 &28.7 &\textbf{19.3} \\
      
      \cdashlinelr{2-12}
&\multirow{2}{*}{$AVG$} 
&AUROC$\uparrow$ &58.9 &60.5 &\textbf{74.6} &57.2 &63.4 &\textbf{74.6} &58.4 &61.2 &\textbf{74.2}   \\ 
&&ECE$\downarrow$	&28.3 &27.3 &\textbf{15.3} &28.9 &26.9 &\textbf{14.1} &31.1 &28.4 &\textbf{15.7}  \\

\midrule
      \multirow{6}{*}{\rotatebox{90}{\textbf{NQ-Open}}} &\multirow{2}{*}{$p(1)$} 
       &AUROC$\uparrow$	&59.4 &62.1 &\textbf{72.3} &55.8 &61.0 &\textbf{72.3} &55.3 &62.8 &\textbf{72.0} \\
      &&ECE$\downarrow$ &30.1 &26.0 &\textbf{17.8} &34.9 &28.7 &\textbf{23.7} &35.1 &29.4 &\textbf{17.5}\\
  
      \cdashlinelr{2-12}
      &\multirow{2}{*}{$p(z-1)$} 
      &AUROC$\uparrow$	&60.4 &57.3 &\textbf{70.9} &57.3 &59.4 &\textbf{67.5} &58.1 &61.3 &\textbf{70.3} \\
      &&ECE$\downarrow$	&29.6 &27.0 &\textbf{20.3} &29.2 &26.3 &\textbf{18.1} &30.4 &30.5 &\textbf{20.5}  \\
   
      \cdashlinelr{2-12}
&\multirow{2}{*}{$AVG$} 
&AUROC$\uparrow$ &60.7 &59.1 &\textbf{75.5} &57.9 &62.3 &\textbf{74.7} &58.8 &64.2 &\textbf{76.9} \\ 
&&ECE$\downarrow$	&27.4 &25.7 &\textbf{14.2} &32.3 &26.1 &\textbf{18.2} &32.8 &28.6 &\textbf{16.4} \\

\bottomrule
\end{tabular}
\end{adjustbox}
     \captionsetup{skip=2pt} % 调整 caption 与正文之间的间距
\label{tab:process_orient}
\end{table*}
\FloatBarrier % 强制结束浮动
\vspace{-2em} % 减少间距

\begin{table*}[t]
\centering
% \footnotesize{
\vspace{-2em} % 减少间距
\caption{\footnotesize{Confidence estimation results across baselines on \textit{Question-oriented} and \textit{Outcome-oriented} tasks.}}

\renewcommand{\arraystretch}{1.1} 
\setlength{\belowcaptionskip}{2pt} 
\renewcommand{\familydefault}{\rmdefault}
\begin{adjustbox}{max width=\textwidth,center}
% {\fontsize{10pt}{12pt}\selectfont 

\begin{tabular}{cc BB dd BB dd BB dd}
    \toprule
     \multirow{2}{*}{\textbf{Models}}  &\multirow{2}{*}{\textbf{Baselines}} &\multicolumn{2}{c}{\textbf{GSM8K}} &\multicolumn{2}{c}{\textbf{CSQA}} &\multicolumn{2}{c}{\textbf{TriviaQA}} &\multicolumn{2}{c}{\textbf{AIME24}} &\multicolumn{2}{c}{\textbf{MMLU}} &\multicolumn{2}{c}{\textbf{NQ-Open}}\\
     % \hline
 % \specialrule{0em}{1pt}{1pt}
 %     \cline{3-4} \cline{5-6} \cline{7-8} \cline{9-10} \cline{11-12} \cline{13-14}

       & & \cellcolor{white}ECE$\downarrow$ & \cellcolor{white}AUROC$\uparrow$  & \cellcolor{white}ECE$\downarrow$ & \cellcolor{white}AUROC$\uparrow$ & \cellcolor{white}ECE$\downarrow$ &\cellcolor{white}AUROC$\uparrow$ 
       & \cellcolor{white}ECE$\downarrow$ & \cellcolor{white}AUROC$\uparrow$
       & \cellcolor{white}ECE$\downarrow$ & \cellcolor{white}AUROC$\uparrow$
       & \cellcolor{white}ECE$\downarrow$ & \cellcolor{white}AUROC$\uparrow$
       \\ 
\midrule
\large
\multirow{7}{*} {\rotatebox{90}{Llama3.1-8B}}
&\textit{P(IK)} &17.6 &72.8 &19.4 &68.7 &20.4 &67.7 &33.1 &67.9 &18.3 &72.1 &22.4 &68.2 \\
&FineCE &13.5 &76.4 &16.0 &68.4 &15.5 &69.8 &18.5 &73.1 &14.3 &76.2 &20.9 &73.1\\
\cdashlinelr{2-14}
&First-Prob &26.2 &66.2 &23.5 &66.8 &24.9 &65.1 &40.3 &65 &21.4 &68.4 &29.4 &66.5\\
&SuC &28.4 &62.0 &32.7 &59.1 &29.7 &60.4 &42.7 &62.2 &24.7 &66.3 &27.3 &61.4\\
&Verb &20.4 &72.9 &28.0 &68.4 &30.1 &69.1 &73.4 &6.1 &31.2 &62.7 &34.0 &65.2\\
&SE &17.6 &73.5 &21.3 &66.7 &19.4 &66.4 &20.9 &68.5 &17.2 &71.2 &22.3 &70.4\\
&FineCE &12.7 &77.1 &14.2 &72.8 &14.6 &70.5 &20.7 &70.4 &12.1 &74.1 &17.1 &75.1\\
\midrule
\multirow{7}{*}{\rotatebox{90}{Qwen2.5-7B}}
&\textit{P(IK)} &17.4 &68.3 &16.3 &68.4 &21.6 &67.9 &27.9 &66.3 &16.1 &69.8 &20.8 &72.3\\
&FineCE &11.4 &72.3 &14.7 &70.6 &15.2 &69.2 &21.2 &76.2 &15.6 &73.1 &17.4 &76.2\\
\cdashlinelr{2-14}

&First-Prob &25.4 &66.4 &26.6 &65.2 &25.9 &62.3 &35.8 &57.4 &30.3 &68.0 &24.5 &68.5\\
&SuC &29.0 &57.4 &28.2 &63.1 &32.7 &58.5 &38.4 &60.4 &27.0 &62.4 &24.1 &63.1 \\
&Verb &15.3 &72.2 &12.4 &70.3 &22.0 &68.4 &78.7 &11.3 &29.4 &63.3 &33.6 &62.4\\
&SE &18.6 &72.1 &19.3 &69.4 &22.5 &68.4 &25.1 &73.5 &22.4 &68.3 &23.8 &71.8\\
&FineCE &10.2 &75.3 &13.1 &70.8 &15.4 &72.5 &17.7 &81.3 &16.3 &75.7 &15.3 &77.8 \\
\midrule
\multirow{7}{*} {\rotatebox{90}{Llama2-13B}}
&\textit{P(IK)} &14.5 &64.8 &29.9 &59.5 &18.7 &65.0 &31.4 &72.1 &17.3 &67.6 &18.3 &70.7 \\
&FineCE &8.9 &67.3 &16.2 &69.3 &15.5 &68.4 &24.8 &78.4 &15.0 &72.6 &13.9 &74.3 \\
\cdashlinelr{2-14}
&First-Prob &23.3 &59.7 &22.3 &60.1 &27.6 &57.1 &42.0 &61.2 &19.4 &64.3 &22.1 &65.1 \\

&SuC &28.8 &57.3 &27.2 &56.7 &23.5 &58.2 &37.3 &57.3 &22.1 &65.2 &24.6 &66.4 \\
&Verb &29.3 &56.2 &21.7 &58.3 &27.1 &53.7 &82.3 &14.9 &32.6 &61.1 &29.8 &62.4 \\
&SE &18.4 &68.6 &16.3 &65.4 &19.5 &63.1 &32.7 &65.1 &20.3 &69.4 &24.1 &70.2 \\
&FineCE &5.1 &77.8 &11.5 &70.5 &12.0 &76.9 &16.2 &75.3 &14.8 &75.4 &14.2 &74.6 \\
\bottomrule
\end{tabular}

\end{adjustbox}

\label{tab:qa_oriented}
\end{table*}

As shown in Table \ref{tab:process_orient}, existing confidence estimation approaches suffer from a fundamental limitation. That is, they fail to capture meaningful uncertainty signals during text generation. FineCE consistently achieves AUROC scores exceeding 70\%, outperforming baseline methods by 10–15 percentage points. In contrast, baselines generally achieve AUROC scores between 57\% and 65\%, indicating performance barely above random chance. Notably, FineCE maintains stable performance across different generation positions (p(1) and p(z-1)), indicating robust confidence estimation throughout the entire generation process. The ECE results further confirm superior calibration, with FineCE achieving significantly lower calibration errors (6.7-16.5\%) compared to baseline methods (19.2-28.3\%).

From Table \ref{tab:qa_oriented}, FineCE consistently outperforms all baselines across both ECE and AUROC metrics on six diverse datasets. The most striking result appears on GSM8K with Llama2-13B, where FineCE achieves an ECE of 5.1\% and AUROC of 77.8\%, representing substantial improvements over the strongest baseline P(IK). This pattern of consistent superiority holds across different model architectures, with FineCE achieving 5-15\% AUROC improvements and 30-60\% relative ECE reductions across experimental conditions.

These results reveal two critical findings. Firstly, existing confidence estimation methods perform poorly across generation positions, often approaching random performance levels. Besides, FineCE's supervised learning method with fine-grained training data construction enables significantly more accurate confidence estimation during the generation. Importantly, these improvements come without sacrificing answer accuracy (the accuracy results are shown in Appendix Table 4), achieved through our replaying strategy and the careful dataset mixing during fine-tuning.

Overall, FineCE consistently enables base models to produce accurate confidence estimates throughout the generation process across diverse tasks, substantially outperforming existing popular confidence estimation methods.
\begin{figure*}[htbp]
  \centering
\includegraphics[width=.99\linewidth]{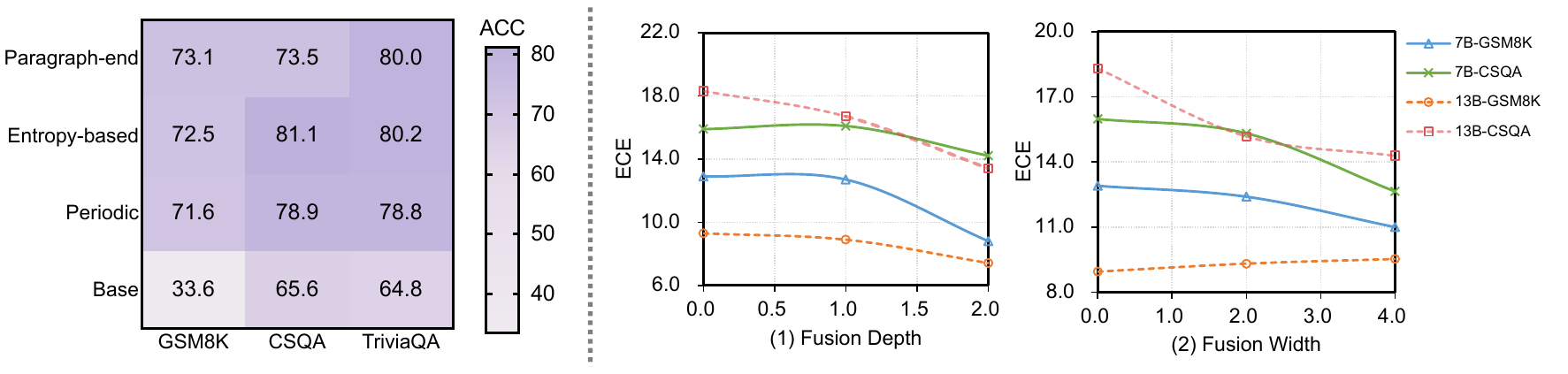}
\caption{{\textbf{(Left:)} Comparison of accuracy between the original model predictions and those selectively accepted by FineCE when the output confidence exceeds 0.8. The backbone used is Llama2-13B. \textbf{(Right:)} Effect of fusion depth (1) and fusion width (2) in FineCE on confidence estimation performance, evaluated with Llama-7B and Llama-13B on the GSM8K and CSQA datasets.
  }
   \label{fig:downstream}
  }
  % \vspace{-2em}
\end{figure*}
\subsection{Downstream Application}
\textbf{RQ2: How does FineCE perform on downstream applications?} 
We evaluate FineCE's practical utility through early-stage confidence estimation and confidence-based filtering. 
From Table \ref{tab:three_strategies}, we observe that \textbf{FineCE achieves reliable confidence estimation using just $\sim$30\% of generated tokens.} Token ratio analysis reveals an interesting pattern: simpler datasets like GSM8K require fewer tokens for reliable estimation (30.4\%), whereas more complex reasoning tasks such as CSQA and TriviaQA require slightly more context (up to $\sim$34\%). This suggests that FineCE adapts its information requirements based on task complexity, with mathematical reasoning allowing earlier confidence assessment than knowledge-intensive or commonsense reasoning tasks.

Furthermore, we implement confidence-based filtering with threshold $\delta$, retaining only responses exceeding the confidence threshold. From Figure \ref{fig:downstream} (Left), we observe FineCE shows consistent accuracy improvements across datasets. The strong correlation between partial-response confidence estimates and final answer correctness validates its effectiveness as a output quality gate, enabling models to reject low-confidence responses before full generation. This capability is particularly valuable in deployment scenarios demanding computational efficiency and reliability, as it enables early termination of potentially incorrect responses.

\begin{table}[!ht]
    \centering
        \caption{Performance comparison of three strategies for identifying optimal calibration positions. \textit{Token Ratio} represents the proportion of  tokens preceding the calibration position relative to the total number of tokens in the complete answer. The backbone model used is Llama2-13B.}
    % \resizebox{0.7\linewidth}{!}{
    \renewcommand{\familydefault}{\rmdefault}
    \begin{adjustbox}{max width=.99\linewidth,center}
    \normalsize
    \begin{tabular}{ccccB}
    \toprule
         Dataset & Strategy  & $ECE_{p_{1}}$ & $ECE_{avg}$ 
         &\multicolumn{1}{c}{Token Ratio}\\ 
         \toprule
        \multirow{3}{*}{GSM8K} & Paragraph  & 9.8 & 7.7 & 30.4\%  \\ 
        &Entropy & 13.2 & 7.7 & 10.0\%  \\
        &Fixed-token & 13.1 & 10.8 & 23.5\%  \\
        \midrule
        \multirow{3}{*}{CSQA} & Paragraph & 26.8 & 13.0 & 22.0\%  \\ 
       &Entropy  & 27.1 & 18.8 & 7.0\%  \\ 
       ~&Fixed-token  & 24.2 & 20.7 & 34.7\%  \\ 
       \midrule
       \multirow{3}{*}{TriviaQA} & Paragraph& 17.2 & 14.5 & 28.5\%  \\
        &Entropy & 18.5 & 15.4 & 13.4\%  \\
       &Fixed-token & 20.0 & 18.0 & 34.1\%  \\ 
        
        % ~ & TriviaQA  & 17.2 & 14.5 & 28.5\%  \\ \midrule
        % \multirow{3}{*}{Entropy} & GSM8K  & 13.2 & 7.7 & 10.0\%  \\ 
        % ~ & CSQA  & 27.1 & 18.8 & 7.0\%  \\ 
        % ~ & TriviaQA  & 18.5 & 15.4 & 13.4\%  \\ \midrule
        %  \multirow{3}{*}{Fixed-token} & GSM8K  & 13.1 & 10.8 & 23.5\%  \\ 
        % ~ & CSQA  & 24.2 & 20.7 & 34.7\%  \\ 
        % ~ & TriviaQA  & 20.0 & 18.0 & 34.1\%  \\ 
        \bottomrule
    \end{tabular}
    \end{adjustbox}
    \label{tab:three_strategies}
\end{table}
\subsection{Further Analysis}
%我们使用Llama-7b分析了FineCE使用三种寻找最佳校验位置策略下的ECE表现情况，结果如表所示。实验设置中，我们的entopy阈值设置为1e-10，固定token策略中，token的长度我们设置为30. 我们发现这三种策略性能表现基本相近,在段落结束之后进行置信度校验的性能相对更好。可能的原因是按照段落进行截断，比较完整的保留了语义的信息，更有利于校验。此外，我们发现使用entrop-based策略在生成过程中会更容易开启校验(ratio占比很小)
\textbf{RQ3: Where does FineCE perform the confidence estimation?}
We conduct a comparative analysis of three calibration position strategies using the Llama2-13B model. For the Entropy-based strategy, we set the entropy threshold to \textit{1e-10}, while for the Periodic Calibration strategy, we fix the calibration interval to every 30 tokens. The results are presented in Table \ref{tab:three_strategies}.

We observe that \textbf{all three strategies demonstrate comparable performance in terms of ECE, with Paragraph-end Calibration strategy yielding slightly better results.} We attribute this improvement to the fact that calibrating at paragraph boundaries helps preserve the full semantic context, which is essential for reliable confidence estimation.

Based on these findings, we draw the following insights. For general tasks, performing confidence estimation at paragraph boundaries is both efficient and effective, significantly reducing unnecessary token consumption. In contrast, for more complex tasks that require finer-grained assessment, the Entropy-based strategy achieves more frequent and accurate confidence estimation through dynamic calibration guided by uncertainty.

% For general tasks, it is sufficient to estimate at the end of paragraph, which alleviate token consumption. 

% For more complex tasks, employing entropy-based strategies for dual verification may be more beneficial, as they enable more frequent and accurate confidence estimation.

% Moreover, the Entropy-based strategy tends to trigger calibration earlier in the generation process, as indicated by smaller ratio values. This suggests that the Entropy-based strategy is more likely to perform confidence estimation more frequently during generation.

% 这意味着在生成过程中，这种策略有很大的可能会频繁执行置信度估计。 因此，我们可以提供基本的准则来辅助找到最佳位置：在一般问题上，按照段落对生成过程进行检测已经足够，而且在某种程度能够减少token消耗；而在更加复杂的任务上，可以使用基于熵的策略来执行双重验证
\textbf{RQ4: How effective is the BCI strategy?} 
We conduct ablation experiments on GSM8K and CSQA datasets using Llama2-7B\footnote{https://huggingface.co/meta-llama/Llama-2-7b} and Llama2-13B models to evaluate the impact of the BCI strategy. Figure \ref{fig:downstream} (Right) shows ECE results for p(1), where d=0 and w=0 represents the FineCE baseline without BCI.

The results demonstrate that BCI consistently improves calibration across all model-dataset combinations. As fusion depth $d$ increases from 0 to 2, ECE drops substantially. On CSQA with Llama2-7B, ECE decreases from 15.3 to 12.6. Similarly, increasing fusion width $w$ from 0 to 4 yields progressive calibration gains, with ECE reductions of up to 15\% on CSQA datasets.

The improvements are particularly pronounced for larger models and more complex reasoning tasks. Llama2-13B benefits more significantly from BCI than Llama2-7B, suggesting that BCI becomes more effective as model capacity increases. Interestingly, CSQA shows greater sensitivity to fusion width compared to GSM8K, indicating that knowledge-intensive tasks require broader cross-attention integration to capture diverse reasoning pathways.
\section{Related Work}
\textbf{Verifier and Calibration Model.}
Although the calibration model and the verifier take similar inputs and produce comparable outputs, they are fundamentally distinct in function. The verifier is designed to assess the quality of a given response in a model-independent manner, assigning consistent evaluation scores regardless of which language model produced the answer \cite{McAleese2024LLMCH, Ke2023CritiqueLLMTA, Huang2024AnES}. In contrast, the calibration model estimates the probability that a specific output is correct, given the behavior of the generating model. This confidence score is inherently model-dependent, as different language models may generate varying responses to the same input, each with different likelihoods of being correct \cite{atil2024llm, Song2024TheGT, renzeEffect}. To sum up, the verifier facilitates a standardized evaluation of generation quality across different models; the calibration model captures model-specific epistemic uncertainty during the generation process, reflecting each model’s unique knowledge confidence. 

% Several studies have explored evaluating reasoning steps \cite{wang2024mathshepherdverifyreinforcellms, lightman2023letsverifystepstep} or final answers \cite{Cobbe2021TrainingVT} by training reward models. These methods aim to rank multiple generated solutions, select the best one, or construct step-wise supervision data \cite{Lai2024StepDPOSP}. However, they are typically tailored to specific tasks such as mathematical reasoning, and provide discrete evaluation scores to enhance final performance. Moreover, they often neglect the accuracy and reliability of the evaluation itself. Our work focuses on developing a generalizable method that delivers fine-grained and accurate confidence estimates for arbitrary text outputs, and and evaluates the calibration capability based on the model's actual responses.

\textbf{Confidence Expression in LLMs.}
Recent studies have explored how LLMs express confidence, mainly through internal signals or explicit verbalization. Leverage internal representations or logits to estimate uncertainty\cite{Su2024UnsupervisedRH, Chen2024SelfIESO, internal}. For example, \cite{Chen2024INSIDELI} analyzes eigenvalues from internal vectors to detect errors, while \cite{robinson2023leveraginglargelanguagemodels} uses token-level logits to measure the uncertainty. Others introduce components like a ``Value Head'' to probe self-assessed confidence \cite{Kadavath2022LanguageM}, but these methods are limited to structured tasks. Another line of work prompts LLMs to verbalize their confidence directly\cite{zhou2023navigating, xiong2023can, Zhang_Yao_Liu_Qin_Wang_Deng_2024}. Techniques include few-shot prompting \cite{Branwen}, supervised training with external labels \cite{tian2023just}, and explicit guidance for confidence output \cite{lin2022teaching}. However, models often exhibit overconfidence and struggle with complex instructions \cite{xiong2023can}.

\section{Conclusion}
In this paper, we propose FineCE, a fine-grained confidence estimation method designed to provide accurate confidence scores throughout the  generation process. We first differentiate FineCE from existing popular confidence estimation approaches, emphasizing its unique advantages. We then detail the training dataset construction procedure used in FineCE, followed by the introduction of three basic strategies to identify the optimal confidence estimation positions. Additionally, during the inference stage, we further present the BCI strategy, which enhances confidence estimation by incorporating the future text to provide a more comprehensive estimation for the current output. Extensive experiments demonstrate that FineCE consistently outperforms existing methods across various confidence estimation tasks. We also validate its effectiveness on several downstream applications.
% In this paper, we propose a fine-grained confidence estimation method FineCE to provide accurate confidence scores throughout the generation process. We begin by highlighting the differences between FineCE and existing popular confidence estimation methods. 
% 接下来详细介绍了FineCE构造训练数据的流程and then describe the dataset construction process. We introduce and three strategies for identifying the optimal estimation position在生成过程中. and the BCI to provide a comprehensive confidence estimate for the current text  在推理阶段。
% Extensive experiments demonstrate the superior performance of our proposed method across various confidence estimation task and downstream application. 
\section{Limitations}
Although FineCE demonstrates effectiveness in providing accurate confidence scores across various confidence estimation task, it encounters challenges when applied to highly open-ended problems, similar to all existing confidence estimation methods. For example, questions like ``\textit{How to stay healthy?}" lack explicit and clear response constraints such as perspective, scope or response length. The inherent ambiguity and broad range of potential solutions in such queries present significant challenges for reliable confidence estimation. We discuss this in detail in the appendix \hyperlink{high-open}{RQ8}. In future work, we will focus on exploring more robust confidence estimation methods specifically tailored to handle these highly open-ended questions.

\bibliography{reference}

\appendix
\clearpage
\section{Appendix}
\label{sec:appendix}
\subsection{Algorithm}
\begin{algorithm}
\small{
\caption{Confidence Estimation Dataset Construction}
\begin{algorithmic}[1]
\Require Generation model $M$, Question set $\mathcal{Q} = \{x_1, x_2, \cdots, x_N\}$, Number of samples $k$, Number of clusters $m$, Number of truncations $\mathcal{T}$
\Ensure Confidence estimation dataset $\mathcal{D} = \{\langle s, \text{Conf}_s \rangle\}$. Initialize $\mathcal{D} \gets \emptyset$
\For{each question $x \in \mathcal{Q}$}
    \State Generate $k$ answers $\{A_x^1, A_x^2, \cdots, A_x^k\}$
    \State Compute confidence score $\text{Conf}_{x}$ based on Equation (2)
    \State Add $\langle x, \text{Conf}_{x} \rangle$ to dataset $\mathcal{D}$
    % 第一次截断
    \State Collect all partial answers $\{A_x^{1*}, \cdots, A_x^{k*}\}$ by truncating \(k\) answers\Comment{{\color{mypink}the first truncation}}
    \State Cluster the partial answers into $m$ clusters $\{C_1, C_2, \cdots, C_m\}$ \Comment{{\color{mypink}cluster only once}}
    \For{$t = 2$ to $\mathcal{T}$}
        \If{$t = 2$}
            \State Select representative centroids from each cluster, $\overline{c}_t \gets \{c_1, c_2, \cdots, c_m\}$
        \Else    \quad $\overline{c}_t \gets \overline{c}^{'} $\Comment{{\color{mypink} partial answers in the $t-1$th truncation }}
        %     \State Select one centroid from $C^{t}$ and denote it as $\overline{c}_t$
        \EndIf
        \State $\overline{c}^{'} \gets \emptyset$\Comment{{\color{mypink}new partial answers}}
        \For{each partial answer $c_i \in \overline{c}_t$}
            \State Concatenate $s_i \gets x \oplus c_i$. Generate $k$ answers based on $s_i$  \Comment{{\color{mypink}completion}}
            \State Compute confidence score $\text{Conf}_{s_i}$ based on Equation (2)
            \State Add $\langle s_i, \text{Conf}_{s_i} \rangle$ to dataset $\mathcal{D}$
            \State Truncate the newly generated $k$ answers \Comment{{\color{mypink}the $t$th truncation}}
            \State Find the semantic centroid $c_{i}^{'} $ among the $k$ truncated results. $\overline{c}^{'}  \gets\overline{c}^{'}  \bigcup \{c_{i}^{'} \} $ \Comment{{\color{mypink}append}}
        \EndFor
    \EndFor

    \For{a complete answer $A$ of question $x$} \Comment{{\color{mypink}confidence score for a complete answer}}
        \If{$A$ is a correct answer} Add $\langle x \oplus A, 1.0 \rangle$ to dataset $\mathcal{D}$
        \Else \quad Add $\langle x \oplus A, 0.0 \rangle$ to dataset $\mathcal{D}$
        \EndIf
    \EndFor
\EndFor

\State \Return $\mathcal{D}$
\end{algorithmic}
}\end{algorithm}
\label{alg:conf_data}

\begin{figure*}[t]
  \centering
\includegraphics[width=.9\linewidth]{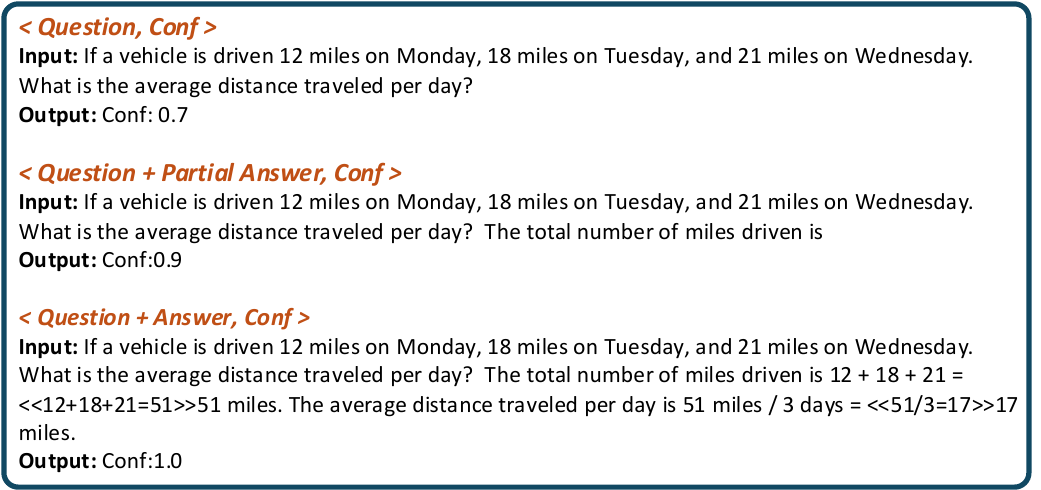}
  \caption{The three types of training data format. }
\label{fig:training_data_format}
\end{figure*}

As discussed in Section \ref{construction}, we provide the algorithmic details of how FineCE employs Monte Carlo sampling to generate three types of data, as illustrated in Algorithm \ref{alg:conf_data}. We also provide three types of training data format in Figure \ref{fig:training_data_format}.
\subsection{Experiments Details}
\label{exp_details}
\subsubsection{Baselines.} 
\begin{figure*}[t]
  \centering
\includegraphics[width=.9\linewidth]{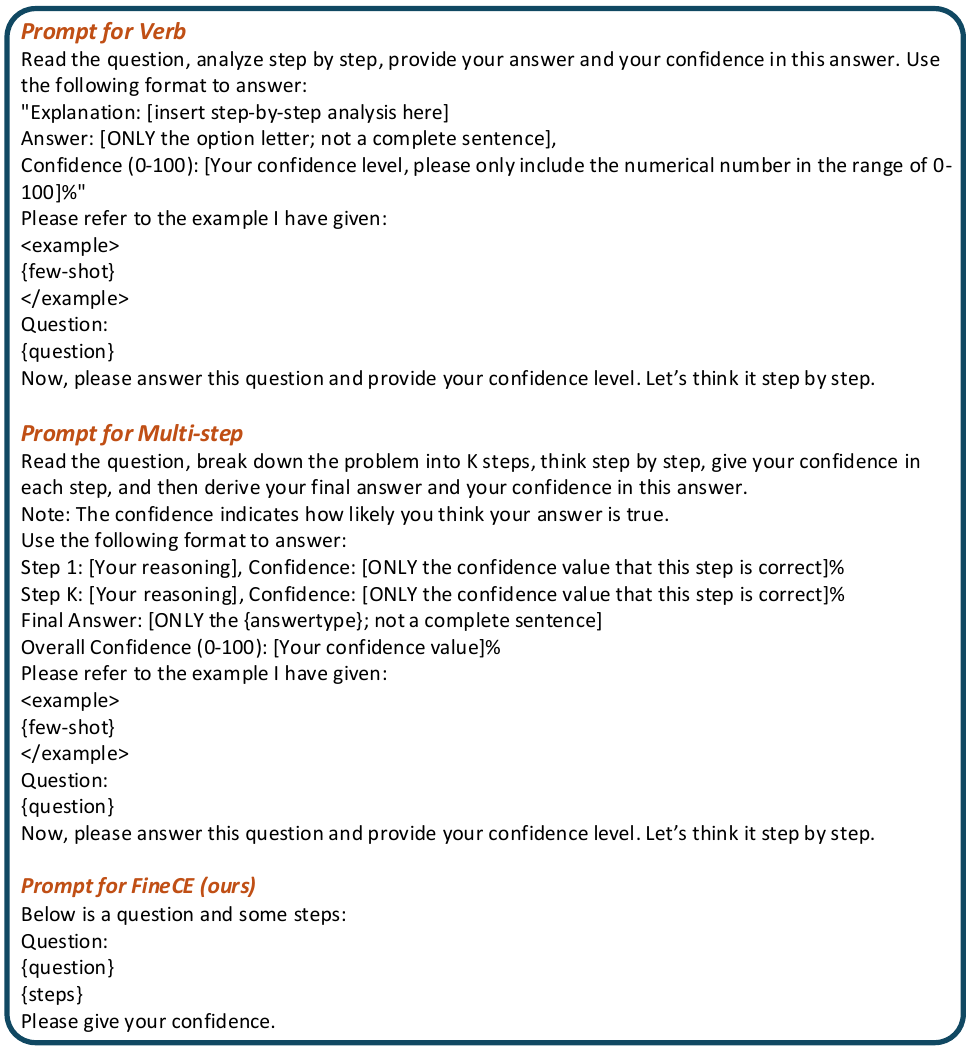}
  \caption{The prompts used in the baselines. }
  \label{prompts}
\end{figure*}

We introduce each method in the baseline, and the prompts used are shown in Figure \ref{prompts}.
\noindent
\paragraph{P(IK).}  It trains a logistic regression with the additional value ``head" added to the model to output the confidence estimated. 
\paragraph{First-Prob.} It uses the logits of the first token of LLM's generated answer as the confidence estimate.
\paragraph{SuC.} It first clusters the sub-questions and uses the same confidence estimate for the questions in the same cluster.
\paragraph{Verb.} It is a prompt-based method. It designs the prompts to guide the model to output its confidence score along with the generated answer. 
\paragraph{LECO.} It also proposes to leverage logits to estimate the confidence of the steps. In addition, it further designs three logit-based scores that comprehensively assess
confidence from both intra- and inter-step perspectives.
\paragraph{Multi-Step.} It also uses prompts to guide the model to output the confidence of the process and takes the average as the final result.

Additionally, we don't use self-consistency as a baseline. While self-consistency has been used in some prior works, we chose not to include it due to two key reasons.

Firstly, \textbf{self-consistency is not a confidence estimation method}. Self-consistency estimates $p(a|q)$, which represents the probability of generating an answer to a given question $q$. Confidence estimation measures are defined as:
    \[
        Conf_s = p(y = \bar{Y} | s, M)
    \]
    (Equation 1), which represents the probability that the predicted answer is correct given the sample and model. Self-consistency conflates generation frequency with correctness probability. A model might consistently generate the same incorrect answer across multiple samples, yielding high self-consistency scores despite being wrong. For example, for the question "1 + 1 = ?", if a model generates "3" in 8 out of 10 samples, self-consistency would assign a confidence score of 0.8. However, this high score doesn't reflect the actual probability that "3" is the correct answer. It merely indicates the model's consistent preference for this response.

The second reason is \textbf{experimental fairness}. Our method and all other baselines operate under single-pass inference. Self-consistency requires multiple forward passes, introducing significant computational overhead and making comparisons unfair.

% \input{training data}

% \label{additional_exp}
% \documentclass{article}
% \usepackage[ruled,vlined]{algorithm2e}
% \usepackage{algorithm}
% \usepackage{algorithmic}
% \usepackage{amsmath}
% \usepackage{lineno}
% \usepackage{booktabs}
% \begin{document}
\begin{table*}[ht]
\renewcommand{\familydefault}{\rmdefault}
\centering
\caption{\footnotesize{Performance of different methods on various benchmarks.}}
\begin{adjustbox}{width=.75\textwidth}
\begin{tabular}{lBdBdBdR}
\toprule
\textbf{Method} & \cellcolor{white}{\textbf{GSM8K}} & \cellcolor{white}{\textbf{CSQA}} & \cellcolor{white}{\textbf{TriviaQA}} & \cellcolor{white}{\textbf{AIME24}} & \cellcolor{white}{\textbf{MMLU}} & \cellcolor{white}{\textbf{NQ\_Open}} & \cellcolor{white}{\textbf{AVG}} \\
\toprule
\rowcolor{bgcolor}\multicolumn{8}{c}{\textbf{Llama3.1-8B}} \\
Base & 72.8 & 78.3 & 74.4 & 13.3 & 55.6 & 50.4 & 57.47 \\
\cdashlinelr{1-8}
P(IK) & 57.4 & 71.0 & 73.3 & \underline{10.0} & 48.4 & 46.1 & 51.0 \\
First-Prob & \textbf{69.4} & \underline{76.4} & \textbf{76.1} & \textbf{13.3} & \underline{53.1} & \textbf{49.3} & \textbf{56.3} \\
SuC & 60.1 & 76.2 & 70.8 & \underline{10.0} & 50.9 & 45.6 & 52.3 \\
FineCE & \underline{61.7} & \textbf{77.4} & \underline{73.9} & \textbf{13.3} & \textbf{54.8} & \underline{48.2} & \underline{54.9} \\
\midrule
\rowcolor{bgcolor}\multicolumn{8}{c}{\textbf{Qwen2.5-7B}} \\
Base & 83.6 & 87.3 & 79.4 & 13.3 & 60.2 & 42.9 & 61.1 \\
\cdashlinelr{1-8}
P(IK) & 70.7 & 77.9 & 73.0 & 13.3 & 54.1 & 40.3 & 54.9 \\
First-Prob & \textbf{79.4} & \underline{80.7} & \textbf{80.2} & \underline{16.7} & \underline{60.2} & \underline{41.4} & \textbf{59.8} \\
SuC & \underline{74.1} & 79.2 & 74.3 & \underline{16.7} & 58.3 & 40.0 & 57.1 \\
FineCE & 73.4 & \textbf{81.1} & \underline{77.3} & \textbf{20.0} & \textbf{60.6} & \textbf{43.6} & \underline{59.3} \\
\midrule
\rowcolor{bgcolor}\multicolumn{8}{c}{\textbf{Llama2-13B}} \\
Base & 31.0 & 64.3 & 65.1 & 3.3 & 43.9 & 41.5 & 41.52 \\
\cdashlinelr{1-8}
P(IK) & 30.4 & \textbf{69.9} & \textbf{66.2} & \underline{0.0} & 38.4 & 35.2 & \underline{40.02} \\
First-Prob & 30.4 & 62.5 & 63.1 & \textbf{3.3} & 39.3 & \underline{39.2} & 39.63 \\
SuC & \underline{31.0} & 60.1 & 62.8 & \underline{0.0} & \underline{40.3} & 37.1 & 38.55 \\
FineCE & \textbf{33.6} & \underline{65.6} & \underline{64.8} & \textbf{3.3} & \textbf{43.1} & \textbf{40.6} & \textbf{41.83} \\
\bottomrule

\end{tabular}
\end{adjustbox}
\label{tab:acc}
\end{table*}

\begin{figure*}[t]
  \centering
  \includegraphics[width=1\textwidth]{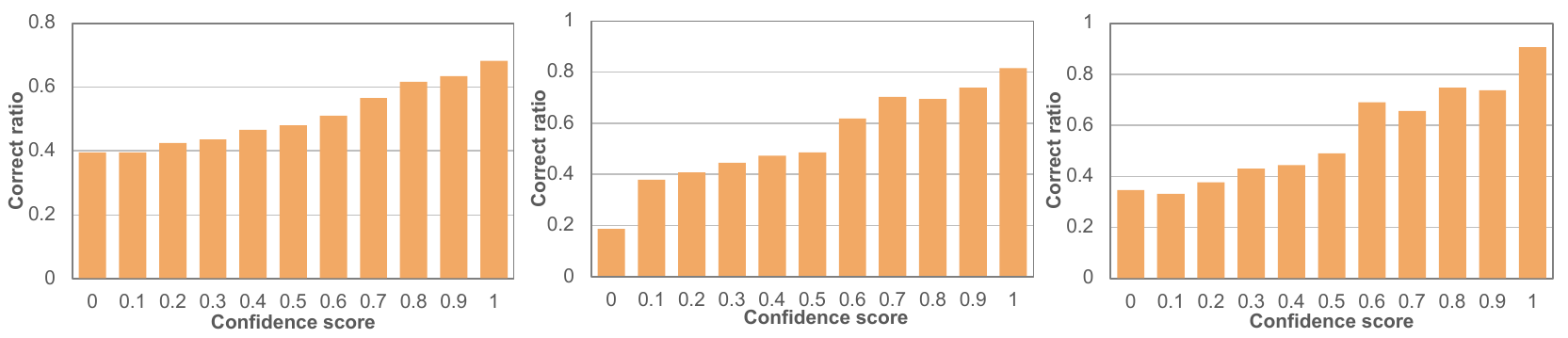}
  \caption{\footnotesize{The Zero-shot performance on OpenBookQA dataset. From left to right, the figures show the confidence estimation performance of FineCE for the question, partial answer, and complete answer. The x-axis represents the confidence scores (\%), and the y-axis represents the ratio of quantities. The top area contains the detailed values of ECE and AUROC.}}
  \label{openbook}
\end{figure*}
\begin{figure*}[ht]
  \centering
  \includegraphics[width=0.8\textwidth]{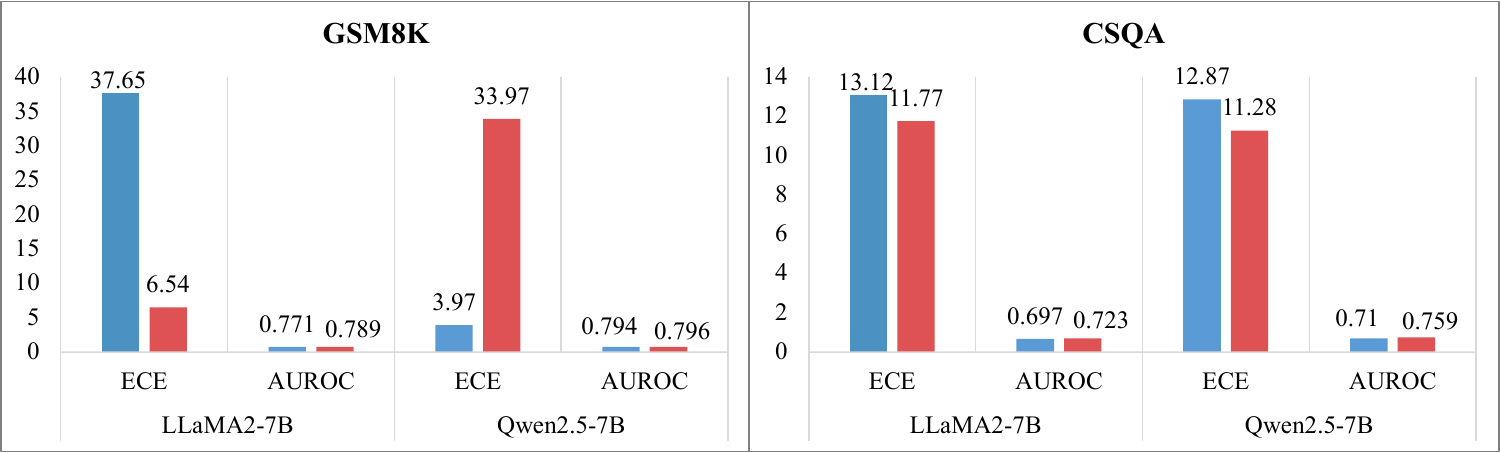}
  % \vspace{e}
  \caption{\footnotesize{On GSM8K(left) and CSQA(right) dataset, the performance confidence estimation for the two different families models using datasets from different sources.The horizontal axis represents the base models.}}
  \label{appendix_all}
  % \vspace{-1em}
\end{figure*}

\begin{figure*}[htbp]
  \centering
\includegraphics[width=.8\linewidth]{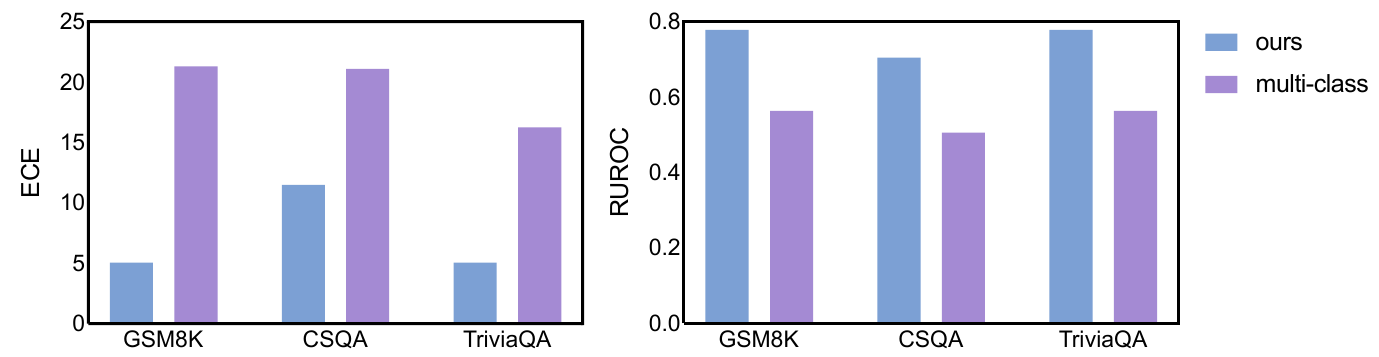}
  \caption{\footnotesize{The performance comparison using different training technical. The backbone model is LLaMA2-13B.}}
  \label{training_skill}
\end{figure*}

\subsubsection{Important Parameters Settings.}
% 构造训练数据时，每一个文本sampling的次数k设置为30
% During fine-tuning, we 实现基于llama-factory。
During training data construction, each text is sampled \(k=30\) times. During the fine-tuning, our implementation is based on  LLaMA-Factory \footnote{https://github.com/hiyouga/LLaMA-Factory}.
We employ the AdamW optimizer with $\beta _{1} = 0.9 $ and $\beta _{2} = 0.5 $. The initial learning rate is set to 1e-4, with the warmup phase of 300 steps. All experiments are conducted on the workstations of NVIDIA A800 PCIe with 80GB memory and the environment of Ubuntu 20.04.6 LTS and torch 2.0.1.

% \textbf{Prompts.} All the prompts used in this paper are shown in Table \ref{prompts}.

\textbf{Accuracy Performance.} The accuracy results are shown in Table 4.

\subsection{Further Discussions}
\label{further_discuss}
\paragraph{RQ5: How does FineCE perform with zero-shot prompt on new task?} To evaluate the generalizability of the FineCE method, we test the confidence estimation performance of FineCE on OpenBookQA dataset \cite{Mihaylov2018CanAS} using Llama2-13B, and the results are shown in Figure \ref{openbook}.

We find that FineCE exhibits outstanding performance across both the ECE and AUROC confidence metrics on OpenBookQA dataset. Additionally, there is \textbf{a robust positive correlation between the model's confidence estimates and the actual accuracy of the answers}. Specifically, we observe that higher confidence levels correlated with higher accuracy. It indicates that our method possesses \textbf{noteworthy generalization capabilities} and is capable to offer reliable confidence estimates when applied to new tasks. 

\begin{figure*}[t]
  \centering
\includegraphics[width=.9\textwidth]{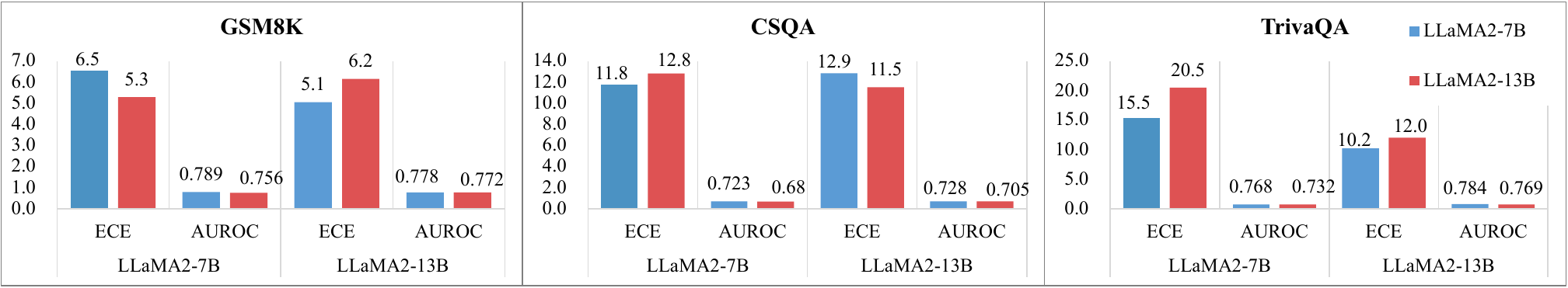}
  \caption{\footnotesize{The performance confidence estimation for two base models using training datasets from different sources. The horizontal axis represents the base models.}}
  \label{training_data}
\end{figure*}

\paragraph{RQ6: How does FineCE perform when trained using datasets from different model?} Here, we use the LLaMA2-13B and LLaMA2-7B as the backbone models. We employ two distinct models to construct the training datasets: the model itself or an alternative model. The results are shown in Figure \ref{training_data}. 

Training with datasets generated from the alternative model achieves confidence calibration performance very close to the obtained using the dataset constructed by the model itself, especially on the GSM8K and CSQA datasets. We guess that it may be related to the used models being from the same family and exhibit significant similarities in their knowledge capabilities. It suggests that larger models could effectively instruct smaller models to learn to express the confidence. In addition, leveraging smaller models to construct training datasets may be a cost-efficient alternative.

% \begin{table}[t]
% \centering
% \caption{\footnotesize{Comparison of the model's accuracy performance across three datasets with a set confidence threshold of 80\%.}}
% \resizebox{.99\linewidth}{!}{
% \begin{tabular}{cccc}
%     \hline
%      Dataset &Base Models &ACC-before &ACC-after\\
%      \hline
%      \multirow{2}{*}{GSM8K} &LLaMA2-7B	 &30.3	 &58.8 (+28.5)\\
%     &LLaMA2-13B	&33.6 &78.3 (+44.7)\\
%     \hline
% \multirow{2}{*}{CSQA} &LLaMA2-7B	&63.7	&79.9 (+16.2) \\
% &LLaMA2-13B	&65.6	&81.8 (+16.2)\\
% \hline
% \multirow{2}{*}{TrivalQA} &LLaMA2-7B	&53.9	&70.3 (+16.4)\\
% &LLaMA2-13B	&64.8	&80.7 (+15.9) \\
% \hline
% \end{tabular}}
% \label{appendix_acc}
% \end{table}

We also use two models from different families to explore this phenomenon further, including Qwen2-7B and LLaMA2-7B, which are from different model families. The results are show in Figure \ref{appendix_all}. We find that there are two different phenomena on different datasets. On the GSM8K dataset, compared with using the model itself to construct training data, the confidence training data constructed with the help of other models performed poorly, especially in the ECE value, where the difference was particularly significant. On the CSQA dataset, the performance difference between the two methods is small. This may be because there is a large difference in the accuracy of Qwen2-7B and LLaMA2-7B on the GSM8K dataset, which makes it impossible to effectively migrate the confidence training data constructed by these two models to each other. 

We can conclude that \textbf{if the performance of two models on a task is close, the confidence training data constructed using one of the models can be effectively used in the training stage of the other model.}

\paragraph{RQ7: Which training skill is more suitable?} On the GSM8K training dataset, we employ two distinct training techniques using the LLaMA2-13B model. One is to add a multi-classification head at the end of the model to output the confidence estimates through classification. The other is the instruction fine-tuning method as we used in the experiment. The outcome confidence estimates results are shown in Figure \ref{training_skill}.

It suggests that \textbf{under the same data scale, the multi-classification techniques exhibited poor performance in confidence estimation task.}

\paragraph{RQ8: How does our method perform on highly open questions?} We randomly select 300 single-round English open question-answering data on Sharegpt $\footnote{\url{https://huggingface.co/datasets/OpenGVLab/ShareGPT-4o}}$, and use LLaMA2-7B to provide confidence estimates. To calculate ECE, we compare the model’s output confidence against the evaluation scores of generated answers obtained from GPT-4. We find that for highly open questions, our proposed method achieved a higher ECE value of 65.66. This is also in line with our expectations. This is because we did not use GPT4's evaluation to assist in constructing training data, resulting in a large difference between the confidence provided by the model and the GPT4 scoring results.
\section{Limitations}
Although FineCE demonstrates effectiveness in providing accurate confidence scores across various confidence estimation task, it encounters challenges when applied to highly open-ended problems, similar to all existing confidence estimation methods. For example, questions like ``\textit{How to stay healthy?}" lack explicit and clear response constraints such as perspective, scope or response length. The inherent ambiguity and broad range of potential solutions in such queries present significant challenges for reliable confidence estimation. We discuss this in detail in the Appendix RQ8. In future work, we will focus on exploring more robust confidence estimation methods specifically tailored to handle these highly open-ended questions.
\end{document}